\documentclass[10pt,twocolumn,letterpaper]{article}

\usepackage{3dv}
\usepackage{times}
\usepackage{epsfig}
\usepackage{graphicx}
\usepackage{amsmath}
\usepackage{amssymb}
\usepackage[percentage]{overpic}
\usepackage[normalem]{ulem}

\usepackage[table,xcdraw,dvipsnames]{xcolor}
\definecolor{LightCyan}{rgb}{0.88,1,1}
\definecolor{LightYellow}{rgb}{1,1,0.7}
\definecolor{LightGreen}{rgb}{0.4, 1, 0.6}

\usepackage[pagebackref=true,breaklinks=true,letterpaper=true,colorlinks,bookmarks=false]{hyperref}

\threedvfinalcopy 

\def\threedvPaperID{153} 

\ifthreedvfinal\fi
\begin{document}

\title{Enhancing self-supervised monocular depth estimation \\with traditional visual odometry}

\author{Lorenzo Andraghetti$^1$ \qquad Panteleimon Myriokefalitakis$^1$ \qquad Pier Luigi Dovesi$^1$ \qquad Belen Luque$^1$\\
Matteo Poggi$^2$ \qquad Alessandro Pieropan$^1$ \qquad Stefano Mattoccia$^2$\vspace{1em}\\
\large $^1$Univrses AB\qquad $^2$University of Bologna\\
\normalsize}

\maketitle

\label{abstract}
\begin{abstract}
Estimating depth from a single image represents an attractive alternative to more traditional approaches leveraging multiple cameras.
In this field, deep learning yielded outstanding results at the cost of needing large amounts of data labeled with precise depth measurements for training. An issue softened by self-supervised approaches leveraging monocular sequences or stereo pairs in place of expensive ground truth depth annotations.
This paper enables to further improve monocular depth estimation by integrating into existing self-supervised networks a geometrical prior. Specifically, we propose a sparsity-invariant autoencoder able to process the output of conventional visual odometry algorithms working in synergy with depth-from-mono networks. Experimental results on the KITTI dataset show that by exploiting the geometrical prior, our proposal: i) outperforms existing approaches in the literature and ii) couples well with both compact and complex depth-from-mono architectures, allowing for its deployment on high-end GPUs as well as on embedded devices (\eg, NVIDIA Jetson TX2).
\end{abstract}
\section{Introduction}
\label{Introduction}

Researchers have tackled the problem of estimating depth from images for decades. Understanding depth allows to interpret the environment in three dimensions and ultimately enabling the construction of 3D maps particularly useful for autonomous navigation of mobile robotics platforms or augmented and virtual reality applications. 

Most of the traditional approaches to estimate depth rely on the assumption of having multiple observations of the scene, either in time (e.g structure from motion) or in space (e.g. stereo or multi-view setup), and exploit hand-crafted features to find correspondences between images to estimate sparse depth measurements of the observed scene \cite{hartley_2003}. 
More recently, machine learning approaches have shown remarkable advances in the field \cite{fu2018supervised}, enabling the dense estimation of depth from a single image, given that a large amount of labelled data is available at training time. 

Self-supervised paradigms relax this constraint \cite{Godard1,Zhou_2017_CVPR}, replacing the need for ground truth depth annotations, usually obtained by means of active sensors \cite{kitti}, with additional images acquired with stereo cameras \cite{Godard1} or a single moving camera \cite{Zhou_2017_CVPR}.
The former strategy is usually more effective, being both the relative pose between the two cameras and the scale factor known.

\begin{figure}
    \centering
    \renewcommand{\tabcolsep}{0.5pt}
    \begin{tabular}{cc}
        \begin{overpic}[width=0.22\textwidth, height=0.11\textwidth]{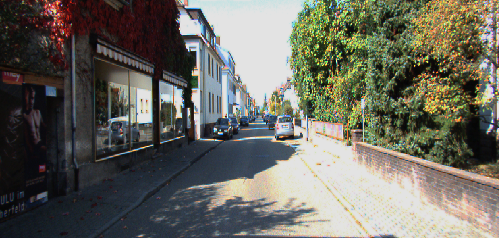}
        \put (2,42) {$\displaystyle\textcolor{white}{\textbf{(a)}}$}
        \end{overpic} 
        & 
        \begin{overpic}[width=0.22\textwidth, height=0.11\textwidth]{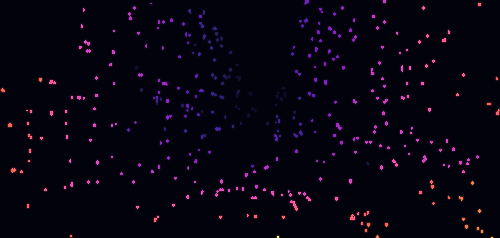}
        \put (2,42) {$\displaystyle\textcolor{white}{\textbf{(b)}}$}
        \end{overpic}
        \\
        \begin{overpic}[width=0.22\textwidth, height=0.11\textwidth]{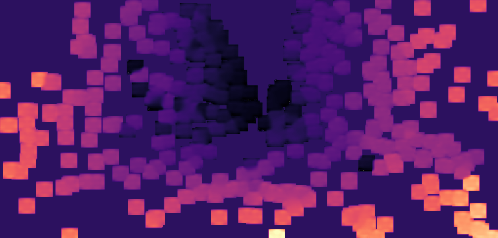}
        \put (2,42) {$\displaystyle\textcolor{white}{\textbf{(c)}}$}
        \end{overpic} 
        & 
        \begin{overpic}[width=0.22\textwidth, height=0.11\textwidth]{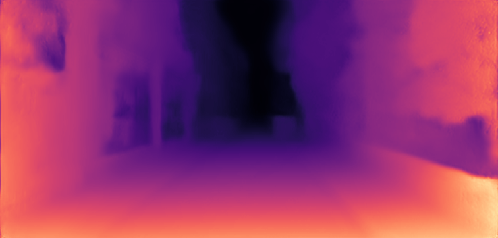}
        \put (2,42) {$\displaystyle\textcolor{white}{\textbf{(d)}}$}
        \end{overpic} \\
    \end{tabular}
    \caption{Monocular depth estimation enhanced by visual odometry (VO). (a) Reference image, (b) sparse 3D points by a monocular VO pipeline, (c) initial densification, (d) final depth map.\label{fig:abstract}}
\end{figure}

Despite the promising results achieved by depth-from-mono frameworks, they often fail in presence of ambiguous environments or elements rarely observed during the training procedure. This is caused by the absence of geometrical cues during the depth predictions, which is mostly learned upon context and semantic content of the observed scene. 
Inspired by the ability of humans in inferring depth from a single eye by leveraging prior knowledge (e.g. the size of known objects)~\cite{howard_2012}, we propose to improve the estimation of depth by introducing a geometrical prior at inference time. Such prior comes in the form of sparse 3D measurements obtained by a traditional visual odometry (VO) algorithm that estimates the structure from consecutive images, most likely scenario occurring in autonomous navigation.

Specifically, we propose a framework that combines a sparsity-invariant \cite{sparse_conv} autoencoder, which enriches our geometrical prior produced by a traditional VO algorithm, with stereo self-supervised models \cite{Godard1,pydnet18} to predict depth-from-mono avoiding the need of ground truth data which is hard to obtain. 
Experimental results on the KITTI dataset \cite{kitti} support the two main contributions of our work:

\begin{itemize}
    \item our framework outperforms self-supervised approaches in literature.
    \item our strategy couples well with both complex \cite{Godard1} and compact \cite{pydnet18} models, making it suited for deployment on high-end GPUs, as well as on embedded devices.
\end{itemize}
We point out that, conversely to traditional depth completion task \cite{sparse_conv} whose aim is to densify an accurate, but sparse set of depth measurements usually sourced by an active sensor such as LiDAR \cite{sparse_conv}, our approach keeps the depth estimation task in the image domain and does not rely on data from any other external source.

Figure \ref{fig:abstract} shows an overview of the proposed approach: given a single image (a) and a set of 3D points obtained through VO (b), these latter are processed by the autoencoder (c) and exploited to support final depth map estimation (d).

\begin{figure*}
    \centering
    \includegraphics[width=0.9\textwidth]{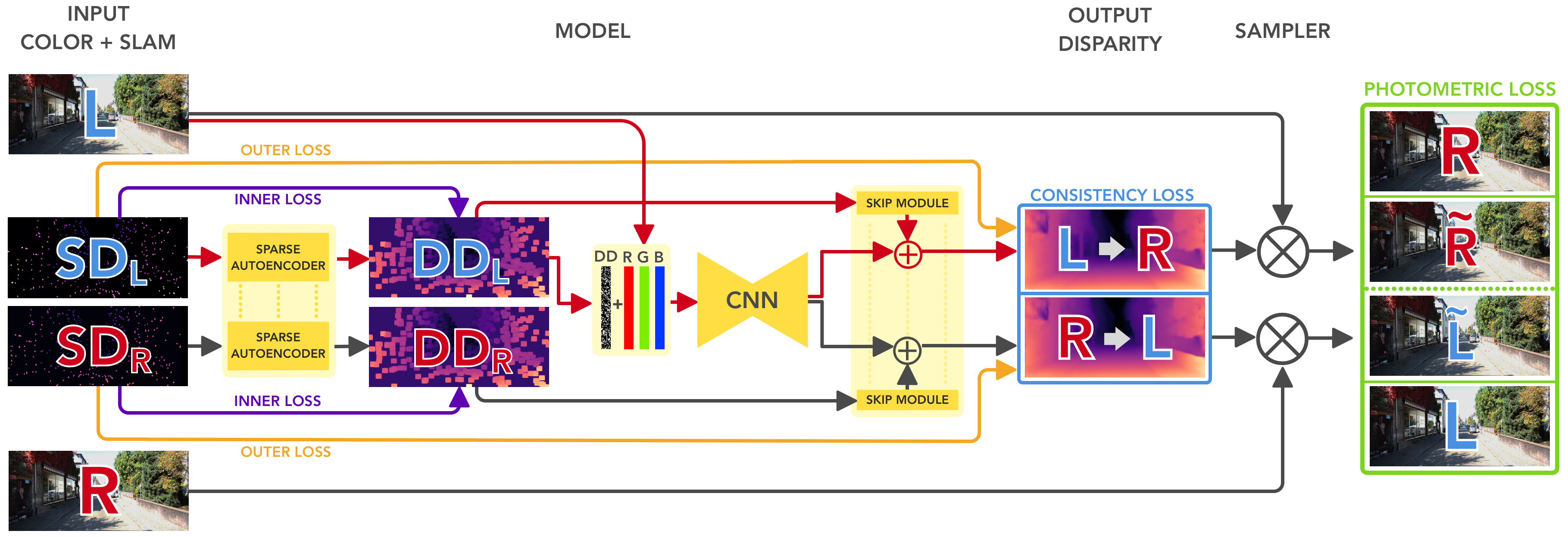}
    \caption{Overview of our framework. Sparse depths (SD) provided by a VO algorithm are fed to a sparse auto-encoder producing more dense depths (DD), forwarded then to the main network together with color image. Self-supervision is obtained by means of stereo image reprojection plus consistency (green and lightblue) and the sparse points themselves (orange and violet). At deployment, monocular cues only are required (red path) \label{fig:model_compare}}
\end{figure*}

\section{Related Work}

We briefly review the literature concerning VO, moving then to
the advances in monocular depth estimation.

\textbf{Visual odometry algorithms.} Large progress has been achieved in the development of VO and SLAM methods \cite{dso,engel14eccv,orbslam2,newcombe2011dtam}.
Although a stereo setup \cite{dso,engel2015large,orbslam2} avoids scale ambiguity, recent trend aims at recovering the scale of monocular VO exploiting geometry \cite{Wang_2018_ICRA,fanani_2017} or deep learning \cite{tateno2017cnn,yin2017scale,yang2018deep}.

Conversely to approaches leveraging depth to improve monocular VO and SLAM \cite{tateno2017cnn,yang2018deep}, in this work we aim at boosting depth-from-mono accuracy exploiting VO.

\textbf{Supervised depth-from-mono.}
The first approaches were supervised and they needed indeed ground truth data to enforce the network to infer depth. Among seminal works, Saxena \etal \cite{Saxena} proposed a method to estimate the absolute scales of different image patches and inferred the depth image using a Markov Random Field model,  
Ladický \etal \cite{Ladicky} incorporated semantic segmentation into their model to improve results.
With the increasing availability of ground truth depth data, supervised approaches based on CNNs \cite{Eigen,liu2016learning} appeared and rapidly outperformed \cite{laina2016deeper,liu2016learning, xu2018supervised} previous techniques. 
State-of-the-art in this field is DORN \cite{fu2018supervised} trained with ordinal regression loss.

\textbf{Self-supervised depth-from-mono}
An attractive trend concerns the possibility of learning depth-from-mono by replacing depth labels with multiple views of the sensed scene and leveraging on image synthesis to obtain supervision signals by having a loss on the reconstructed image.
In general, acquiring images from a stereo camera enables a more effective training than using a single, moving camera, since the pose between frames is known.
Concerning stereo supervision, Garg \etal \cite{Garg} first followed this approach, while Godard \etal \cite{Godard1} introduced a left-right consistency loss. Other methods improved efficiency \cite{pydnet18}, deploying a pyramidal architecture, and accuracy by simulating a trinocular setup \cite{3net18}, including joint semantic segmentation \cite{ramirez2018} or adding adversarial term \cite{Aleotti_monogan_2018,kumar2018gan}. In \cite{DATE_2019}, a strategy was proposed to reduce further the energy efficiency of \cite{pydnet18} leveraging fixed-point quantization. 
In \cite{Pilzer_2019_CVPR} knowledge distillation and cycle consistency proved to improve results, while \cite{Tosi_2019_CVPR} introduces a stacked architecture, namely MonoResMatch, embodying virtual view synthesis and disparity computation and additional \textit{proxy}-supervision self-sourced by running a traditional stereo algorithm \cite{hirschmuller2008stereo}.
Concerning supervision from single camera sequences, Zhou \etal \cite{Zhou_2017_CVPR} were the first to follow this direction. Their approach was improved including additional cues such as point-cloud alignment \cite{Mahjourian_2018_CVPR}, differentiable Direct Visual Odometry (DVO) \cite{Wang_2018_CVPR} and optical flow \cite{zou2018dfnet,Yin_2018_CVPR}. As for stereo supervision, traditional structure-from-motion algorithms (SFM) have been used to provide additional supervision \cite{Klodt_2018_ECCV}. More recently, Casser \etal \cite{casser2019struct2depth} introduced moving object segmentation and online refinement.
Finally, few works combined the best of the two worlds, as in \cite{Zhan_2018_CVPR}. In particular, Yang \etal \cite{yang2018deep} proposed Deep Virtual Stereo Odometry (DVSO), a framework for monocular depth and ego-motion estimation trained on proxy-labels obtained from a stereo odometry algorithm.
Finally, some approaches combine self-supervision with ground-truth labels from either LiDAR \cite{Kuznietsov} or synthetic datasets \cite{luo2018supervised,guo2018learning,atapour2018real}.

As proven in prior works \cite{Tosi_2019_CVPR,Klodt_2018_ECCV,yang2018deep}, we believe that traditional knowledge can provide additional cues to learning-based frameworks for monocular depth estimation. On the other hand, while existing works leverage such knowledge at training time only, we deploy a monocular VO algorithm to obtain geometrical priors to feed our network with. Being such priors sourced by a monocular setup, they are available at inference time in contrast to others available from stereo images  \cite{Tosi_2019_CVPR,yang2018deep} and thus available at training time only.

\textbf{Depth completion.} This category covers a collection of methods with a variety of different input modalities (\eg, relatively dense depth input \cite{Shen_2013_ICCV_Workshops} vs sparse depth measurements \cite{Ma2017SparseToDense}; with color images for guidance \cite{Ma2017SparseToDense} vs. without \cite{sparse_conv}). 
The completion problem becomes particularly challenging when the input depth image has very low density, because the inverse problem is ill-posed. One of the most popular scenario concerns with the use of 3D LiDARs, providing roughly 5\% pixels when reprojected on images \cite{sparse_conv}. Specifically, Ma and Karaman \cite{Ma2017SparseToDense} proposed an end-to-end deep regression model for depth completion. Uhrig \etal \cite{sparse_conv} proposed sparse convolution, a variant
of regular convolution operations capable of dealing with data sparsity in neural networks. Eldesokey \etal \cite{eldesokey2018propagating} improved the normalized convolution for confidence propagation. Chodosh \etal \cite{Chodosh_ACCV_2018} incorporated the traditional dictionary learning with deep learning into a single framework for depth completion. All learning based methods \cite{Ma2017SparseToDense,sparse_conv,eldesokey2018propagating,Chodosh_ACCV_2018} are trained in a supervised manner deploying depth labels.

In contrast we leverage depth estimates obtained by means of a VO algorithm, distinguishing our approach from reviewed depth completion models usually exploiting LiDAR points that are i) sourced from very accurate, active sensors and ii) have an average density of 5\% with respect to the entire image, while the VO pipeline used in our experiments only provides about 0.06\% (\ie 1 every 1600+ pixels).
\section{Method}
\label{Method}

In this section, we introduce the rationale behind the proposed method and the modules deployed in our framework.
We argue that it is unlikely that the entire life-cycle of an application is constrained, on most cases, to a single image acquisition at a single time frame.
A popular example is represented by the autonomous driving scenario, where continuous image acquisition by means of a single camera is necessary.
We aim at improving monocular depth estimation by leveraging this assumption in order to recover the geometry that is missing from a single image acquisition. 
For this purpose, we choose traditional VO algorithms to obtain a set of 3D points used as additional input to our framework to guide it towards more accurate estimations. In particular, sparse 3D points are mapped to image pixels and converted to an equivalent representation with respect to the one of the final depth output. For instance, in case of stereo self-supervision \cite{Godard1} 3D points' depth is back-triangulated to disparity according to the specific setup (\ie, baseline and focal length) deployed for training.
Figure \ref{fig:model_compare} shows our pipeline, made of two main modules: a sparsity-invariant autoencoder, processing the aforementioned 3D points to obtain more dense priors, and a depth estimator that outputs the final depth map when fed with the reference image and densified priors. While stereo images are required at training time, only the monocular input is processed at deployment (connected by the red path in the figure).
In order to provide meaningful information to the network, the input cues are scale-aware. This can be easily obtained at training time by running a stereo VO algorithm \cite{orbslam2}, while at test time a monocular VO framework with any scale recovery, as for instance \cite{Wang_2018_ICRA,fanani_2017}, is required.

\subsection{Sparsity-invariant autoencoder}
\label{sec:autoencoder}

\begin{figure}
    \centering
    \includegraphics[width=0.35\textwidth]{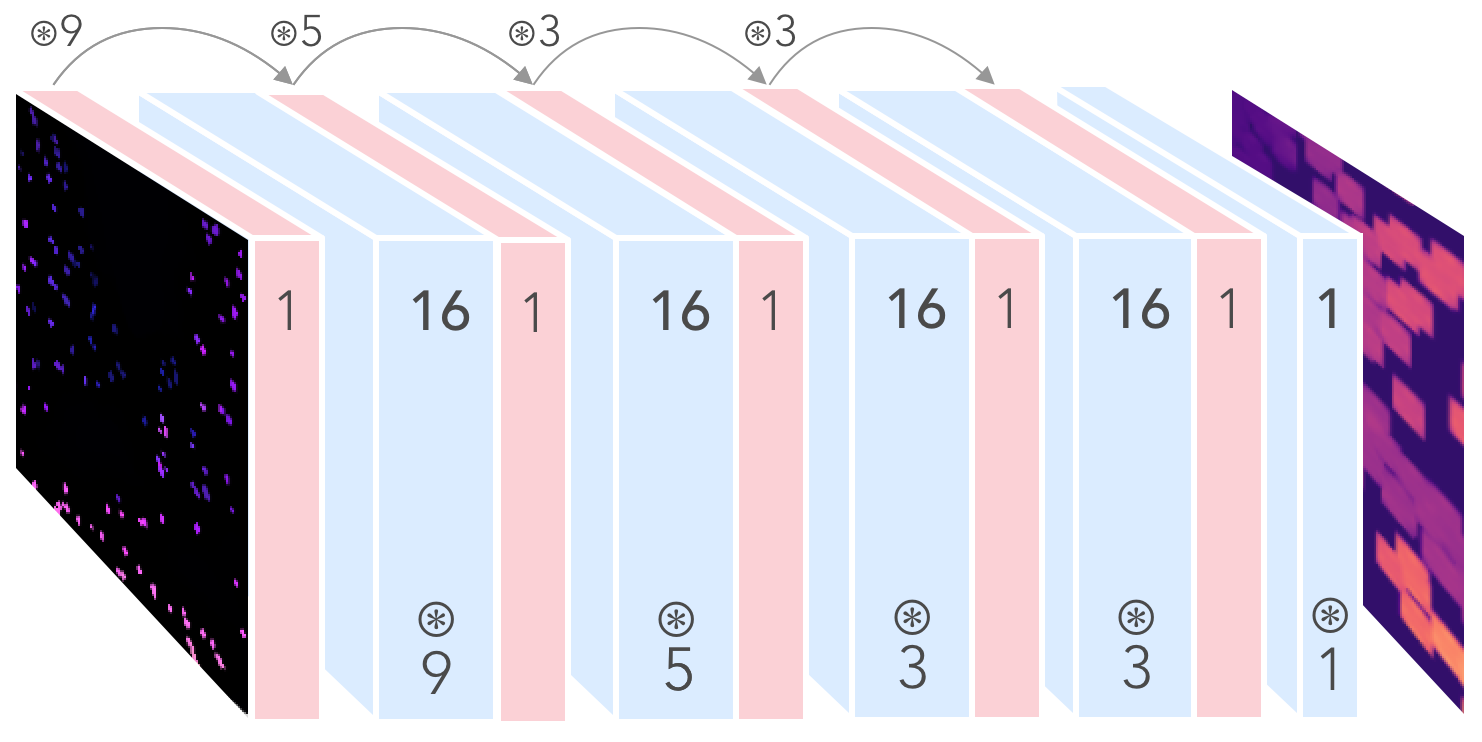}
    \caption{Structure of the sparsity-invariant autoencoder. Four convolutional layers extract 16 features each, respectively with $9\times9, 5\times5, 3\times3$ and $3\times3$ kernels.
    \label{fig:autoencoder}}
    \vspace{-0.3cm}
\end{figure}

The first step in our pipeline consists in processing the 3D points retrieved by means of VO. Because of their sparse nature, we design a sparsity-invariant autoencoder, since traditional convolutions results in poor performance when dealing with such kind of data, as proven in \cite{sparse_conv}.
As shown in Figure \ref{fig:model_compare}, our autoencoder obtains a more dense depth map by means of five sparsity-invariant convolutional layers. The output of this module, namely $DD$ in the figure, is supervised by the \textit{inner} loss shown in the figure and better described in the remainder. Figure \ref{fig:autoencoder} shows how the autoencoder is composed: 4 sparse-convolution layers with decreasing kernel size (9, 5, 3, 3), each one with 16 filters and stride fixed to 1 in order to keep the same resolution of the input. One final sparse-convolution pixel-wise filter is added in order to get an image that represents a denser disparity map which is used for the inner loss. Then, it is concatenated to the input image and forwarded both to the main depth estimator and to a skip residual module that will be further discussed.
Since the output of the VO system is a set of sparse 3D points, it is possible to reproject them onto both left and right camera planes, generating sparse disparity maps $SD_L$ and $SD_R$. During training we employ two autoencoders with shared weights to generate both $DD_L$ and $DD_R$ from left and right sparse ones. Therefore, we enforce consistency in the losses keeping the whole system symmetric. The rationale behind this choice will be discussed shortly, while ablation experiments will highlight the contribution introduced by such strategy.
This symmetry is employed during training only, while at test time a single autoencoder processes left sparse disparity map $SD_L$ to generate $DD_L$ which is given to the depth estimator after a concatenation with RGB color image.

\subsection{Depth estimator}

The recent literature provides multiple architectures for self-supervised monocular depth estimation \cite{Godard1,pydnet18,Tosi_2019_CVPR}.
To prove that our proposal is compatible with both complex and compact networks, making it suited for a wide range of applications on both high-end GPUs as well as on low-power devices, we choose two main models for our experiments: monodepth by Godard \etal \cite{Godard1} and PyD-Net by Poggi \etal \cite{pydnet18}. The main difference between the two consists into the backbone used for features extraction.

The former represents the first model proposed for self-supervised monocular depth estimation from stereo images. In its more accurate version, it consists of a ResNet50 \cite{He_2016_CVPR} encoder and about 58 million parameters. The latter deploys a compact, pyramidal features extractor, counting fewer than 2 million parameters and dramatically reducing both memory and runtime requirements \cite{pydnet18}. Experimental results will show how the proposed pipeline is compatible with different architectures designed to maximize either accuracy \cite{Godard1} or efficiency \cite{pydnet18}.

In Figure \ref{fig:model_compare}, the aforementioned architecture (yellow block) is fed with an RGB image and densified depth cues and outputs two inverse depth maps $D^{LR}$ and $D^{RL}$, \ie disparity maps, aligned respectively with the input image (left frame of a training stereo pair) and the additional one used for supervision (right frame). 

\subsection{Skip module}
In order to lighten the estimation task of the depth estimator, we add a residual skip module, further processing $DD$. 
This module is made of a single ResNet block \cite{He_2016_CVPR}, built by three layers respectively with kernels $1\times1, 3\times3, 1\times1$ and extracting 16, 16 and 64 features. In parallel, a skip connection made by a single $1\times1$ layer produces 64 features summed to those extracted by the latter of the previous three layers. A final $1\times1$ layer produces a residual correction $DD'$.
For symmetry, both $DD_L, DD_R$ are processed by a shared skip module to obtain $DD'_L, DD'_R$.

Finally, we obtain two maps $d^L$ and $d^R$ as last output, respectively summing $D^{LR}$ to $DD'_L$ and $D^{RL}$ to $DD'_R$.
An \textit{outer} loss is computed between these final outputs and SD making the depth estimator, in other words, focusing on the remaining portions of the image for which no prior depth is available. 
Since both $d^L$ and $d^R$ are optimized, the symmetry kept by the autoencoder avoids unbalancing between losses computed on the two, as explained in Section \ref{sec:autoencoder}. 

\subsection{Training Loss}

Following successful attempts in literature \cite{Godard1,pydnet18,Tosi_2019_CVPR}, we deploy a multi-component loss function defined as 

\begin{equation}
	\mathcal{L} = \alpha_{st}\sum_{s=1}^4\mathcal{L}^s_{st} + \alpha_{in} \mathcal{L}_{in} + \alpha_{out} \mathcal{L}_{out}
	\label{eq:monodepth-loss}
\end{equation}
where $\mathcal{L}_{st}$, $\mathcal{L}_{in}$ and $\mathcal{L}_{out}$ are respectively stereo self-supervision, inner and outer losses.

\subsubsection{Stereo self-supervision}

We train our network using stereo self-supervision \cite{Garg}. At training time, each sample consists of a stereo pair made of L and R, respectively, the image input to the model and the one used for image reprojection and the subsequent loss computation. According to Equation \ref{eq:monodepth-loss}, at each scale we compute $\mathcal{L}_{st}$ as

\begin{equation}
\begin{split}
	\mathcal{L}_{st} = &\beta_{ap}(\mathcal{L}_{ap}^L+\mathcal{L}_{ap}^R)+\beta_{ds}(\mathcal{L}_{ds}^L+\mathcal{L}_{ds}^R) \\ +&\beta_{lr}(\mathcal{L}_{lr}^L+\mathcal{L}_{lr}^R) + \beta_{o}(\mathcal{L}_{occ}^L+\mathcal{L}_{occ}^R)
	\label{eq:stereo-loss}
\end{split}
\end{equation}
respectively made of appearance matching, disparity smoothness, left-right consistency and occlusion terms.

\textbf{Appearance Matching Loss.}
enforces the reconstructed image to appear similar to the corresponding training input, combination of L1 and single scale Structured Similarity Index Measure (SSIM) \cite{wang2004image} which compares, for each pixel of coordinates $(i,j)$, the input image $I^L$ and its reprojected $\widetilde{I}^L$ obtained by means of bilinear warping according to disparity estimations.

\begin{equation}
	\mathcal{L}_{ap}^L = \frac{1}{N}\sum_{ij}\gamma\frac{1-SSIM(I_{ij}^L,\widetilde{I}_{ij}^L)}{2}+(1-\gamma)||I_{ij}^L-\widetilde{I}_{ij}^L||
\end{equation}
where $N$ is the number of pixels and $\gamma = 0.85$.

\textbf{Disparity Smoothness Loss}
enforces smooth disparities exploiting an L1 penalty on the disparity gradients $\partial d$, weighted by an edge aware term from the image.

\begin{equation}
	\mathcal{L}_{ds}^L = \frac{1}{N}\sum_{ij}|\partial_x d^L_{ij}|e^{-||\partial_x I_{ij}^L||}+|\partial_y d^L_{ij}|e^{-||\partial_y I_{ij}^L||}
\end{equation}

\textbf{Left-Right Consistency Loss}
enforces the left and the right disparities to be consistent by using an L1 penalty between the left-to-right disparity map and the reconstructed one which comes from sampling the right-to-left in a similar manners as for the left and right images:
\begin{equation}
	\mathcal{L}_{lr}^L = \frac{1}{N}\sum_{ij}|d^L_{ij}- d^R_{ij+d^L_{ij}}|
\end{equation} 

\textbf{Occlusion Loss}
discourages artifacts near occlusions \cite{yang2018deep} by minimizing the sum of all disparities

\begin{equation}
    \mathcal{L}^L_{occ} = \frac{1}{N}\sum_{ij}d^L_{ij}
\end{equation}

\subsubsection{Inner loss}

The purpose of sparsity-invariant autoencoder is to provide the depth estimator with more dense depth priors. To this aim, we enforce the output map $DD$ to be consistent with the input cues $SD$ where these are defined 

\begin{equation}
	\mathcal{L}_{in} = \frac{1}{N}\sum_{ij}|DD_{{ij}} - SD_{L_{ij}}|
\end{equation}
For symmetry, this is carried out on both $DD_L$ and $DD_R$.

\subsubsection{Outer loss}

The final prediction $d^L$ by our network is a sum of $DD'_L$ produced by the skip module and $D^{LR}$ by the depth estimator. Again, since we want to preserve the information sourced by VO, we apply a second, outer loss to enforce consistency between $SD$ and the final output

\begin{equation}
	\mathcal{L}_{out} = \frac{1}{N}\sum_{ij}|d_{ij} - SD_{{ij}}|
\end{equation}
As for the inner loss, this is carried out on both $d^L$ and $d^R$ as well for symmetry.

\section{Experimental results}
\label{results}

In this section, we describe the dataset and the implementation details, and report results concerning our framework in various training/testing configurations, showing that our approach is consistently beneficial to traditional self-supervised approaches. To conform to the literature, we assess the performance of monocular depth estimation techniques following the protocol by Eigen \etal \cite{Eigen}, extracting data from the KITTI \cite{kitti} dataset and using sparse LiDAR measurements as ground truth for evaluation. 

\begin{table*}[!htbp]
\centering
\scalebox{0.95}{
\begin{tabular}{l|cccc|ccc}
&\multicolumn{4}{c}{\cellcolor{blue!25} Lower is better}
 & \multicolumn{3}{c}{\cellcolor{LightCyan} Higher is better} \\
\hline
Model & \cellcolor{blue!25} Abs Rel & \cellcolor{blue!25} Sq Rel & \cellcolor{blue!25} RMSE & \cellcolor{blue!25} RMSE log &  \cellcolor{LightCyan}$\delta<$1.25 &  \cellcolor{LightCyan}$\delta<1.25^2$ & \cellcolor{LightCyan}$\delta<1.25^3$ \\
\hline
Monodepth - ResNet \cite{Godard1} & 0.108 & 0.679 & 4.123 & 0.194 & 0.868 & 0.952 & 0.978 \\
\hline
baseline & 0.109 & 0.660 & 4.077 & 0.195 & 0.866 & 0.952 & 0.979 \\
+ sparse-autoencoder  &  0.099 & 0.666 &  3.910 &  0.200 &  0.878 &  0.947 &  0.973 \\    
+ sparse-autoencoder + skip  &  0.099 & 0.654 &  3.843 &  0.200 &  0.879 &  0.948 & 0.973 \\
+ sparse-autoencoder + skip + sym.  &  0.095 & 0.621 &  3.827 &  0.186 &  0.885 &  0.952 & 0.977 \\
+ sparse-autoencoder + skip + sym. + ft  & \textbf{0.091} & \textbf{0.548} & \textbf{3.690} & \textbf{0.181} & \textbf{0.892} & \textbf{0.956} & \textbf{0.979} \\
\hline
\end{tabular}
}
\smallskip
\caption{Ablation experiments for VOMonodepth on KITTI \cite{kitti} odometry sequences from the Eigen split \cite{Eigen} (8691 frames).}
\label{table:ablation}
\end{table*}

\subsection{Datasets}

For all our experiments we compute standard metrics \cite{Eigen} in the field of monocular depth estimation. Abs rel, Sq rel, RMSE and RMSE log
represent error measures, while $\delta<\varepsilon$ represents the percentage of predictions whose maximum between ratio and inverse ratio with respect to the ground truth is lower than $\varepsilon$, traditionally set to $1.25, 1.25^2$ and $1.25^3$.

For our experiments we use the KITTI dataset \cite{kitti}. It consists of about 42382 rectified stereo pairs grouped into 61 sequences, with an image resolution of $1242\times375$. During acquisition, a LiDAR sensor gathered sparse depth measurements. For our experiments, we divide the entire dataset into a training and a testing set, following the traditional split by Eigen \etal. \cite{Eigen}. In particular, since our method is coupled with VO algorithms, we define different subdivisions, still compliant with the Eigen split:

\begin{itemize}
    \item for training, we adopt the $K_r, K_o$ sets introduced by Yang \etal \cite{yang2018deep}, being the latter part of sequences 01, 02, 06, 08, 09 and 10 from KITTI odometry dataset.
    
    \item for testing, we adopt sequences 00, 04, 05 and 07 from KITTI odometry, that partially overlaps with the Eigen testing set. 
\end{itemize}
This split allows for full deployment of VO algorithms both at training time, described in detail in the remainder, as well as for evaluation.
Focusing on the testing split, the one we introduce counts 8691 frames, in contrast with the original one by Eigen \etal \cite{Eigen} counting only 697 images, yet being fully consistent with it (\ie, there is no overlap between the 8691 frames with any of the frames from the Eigen original training set). This allows for a fair comparison with any method proposed in literature, if trained on the Eigen split and whose weights are provided by the authors, without the need for retraining.

\subsection{Implementation details}

Our framework is implemented using the Tensorflow library. 
We designed two variants, namely \textbf{VOMonodepth} and \textbf{VOPyD-Net}, respectively built around the depth estimators proposed by Godard \etal \cite{Godard1} and Poggi \etal \cite{pydnet18}. 
In both cases, at training time we fed the network with batches of 8 images, using Adam Optimizer \cite{adam} with $\beta_1=0.9, \beta_2=0.999 $ and $\epsilon=10^{-8}$ and a learning rate of $\lambda=10^{-4}$, halved twice after $\frac{3}{5}$ and $\frac{4}{5}$ of the total epochs.  
The weights in our loss function have been tuned respectively to $\alpha_{st}=1$, $\alpha_{in}=5$, $\alpha_{out}=2$, $\beta_{app}=1$, $\alpha_{lr}=1$, $\alpha_{occ}=0.01$ and  $\beta_{ds}=0.1/r$, being $r$ the downsampling factor of each scale.
According to the chosen depth estimator, we run different training schedules: for VOPyD-Net, we run 200 epochs halving the learning rate at 120 and 160, while for VOMonodepth we run 50 epochs and halve at 30 and 40, following in both cases the guidelines suggested by the authors of PyD-Net and Monodepth respectively. We perform established data augmentation procedures \cite{Godard1}, consisting of horizontal flip of the input and color augmentation with with a 50\% chance, random gamma, brightness and color shifts.
At inference time, the same post-processing from \cite{Godard1} is applied by VOMonodepth. Images are resized to $256\times512$ and VO points are reprojected accordingly, except for VOPyD-Net where SD maps are provided at half the resolution for the sake of speed, then estimated DD are upsampled to the original resolution.

In our experiments, we deploy two VO odometry algorithms for training. The two respectively exploit stereo and monocular sequences. While the former provides the correct scale, the second requires a scale recovery mechanisms as for instance in \cite{Wang_2018_ICRA,fanani_2017}. In order to ease the learning process, we perform a first round of training on $K_r + K_o$ feeding the network with the 3D points from stereo VO. Then, we run a further round on $K_o$ switching to the monocular VO used at testing time.
We will show in the experiments how this strategy is beneficial to our framework.

For stereo VO we use ORB-SLAM2 \cite{orbslam2}. As monocular VO algorithm with scale recovery, we use the pipeline developed by Zenuity\footnote{\url{https://www.zenuity.com/}} by their kind concession, the same is deployed for inference in our evaluation keeping our method fully monocular at deployment.

\begin{table*}[!htbp]
\centering
\scalebox{0.88}{
\begin{tabular}{l|c|c|c|cccc|ccc}
\multicolumn{4}{c}{} &\multicolumn{4}{c}{\cellcolor{blue!25} Lower is better}
 & \multicolumn{3}{c}{\cellcolor{LightCyan} Higher is better} \\
\hline
Method & Supervision & PP & VO & \cellcolor{blue!25} Abs Rel & \cellcolor{blue!25} Sq Rel & \cellcolor{blue!25} RMSE & \cellcolor{blue!25} RMSE log &  \cellcolor{LightCyan}$\delta<$1.25 &  \cellcolor{LightCyan}$\delta<1.25^2$ & \cellcolor{LightCyan}$\delta<1.25^3$ \\
\hline
SfmLearner \cite{Zhou_2017_CVPR} & Mono$^*$&  &  & 0.175&1.309&5.515&0.247&0.740&0.916&0.971 \\
Vid2depth \cite{Mahjourian_2018_CVPR} & Mono$^*$&  &  & 0.143&0.827&4.702&0.213&0.812&0.943&0.979 \\
GeoNet - ResNet \cite{Yin_2018_CVPR} & Mono$^*$&  &  & 0.141&0.842&4.688&0.209&0.812&0.945&0.981 \\
LKVO \cite{Wang_2018_CVPR} & Mono$^*$&  &  & 0.135&0.812&4.246&0.200&0.836&0.952&0.982 \\
DF-Net \cite{zou2018dfnet} & Mono$^*$&  &  & 0.131&0.706&4.365&0.196&0.831&0.952&0.984 \\
Struct2depth (M) \cite{casser2019struct2depth} & Mono$^*$&  &  & 0.135&0.792&4.356&0.197&0.836&0.955&0.984 \\
\hline
PyD-Net \cite{pydnet18} & Stereo & &  & 0.130 & 0.833 & 4.569 & 0.219 & 0.825 & 0.938 & 0.974 \\
VOPyD-Net & Stereo & & \checkmark & 0.105 & 0.916 & 4.916 & 0.203 & 0.874 & 0.946 & 0.974 \\
PyD-Net \cite{pydnet18} & Stereo & \checkmark &  & 0.123 & 0.733 & 4.333 & 0.210 & 0.834 & 0.943 & 0.976
\\
VOPyD-Net & Stereo & \checkmark & \checkmark & \textbf{0.102} & \textbf{0.611} & \textbf{3.810} & \textbf{0.188} & \textbf{0.876} & \textbf{0.952} & \textbf{0.979} \\
\hline
Monodepth - ResNet \cite{Godard1} & Stereo & \checkmark & &  0.108&0.679&4.123&0.194&0.868&0.952&0.978 \\
3Net - ResNet \cite{3net18} & Stereo & \checkmark & & 0.106 & 0.627 & 3.982 & 0.192 & 0.869 & 0.953 & 0.979 \\ 
MonoResMatch \cite{Tosi_2019_CVPR} & Stereo+SGM & \checkmark & & 0.102 & 0.563 & 3.725 & 0.183 & 0.885 & \textbf{0.964} & \textbf{0.986} \\
VOMonodepth - ResNet & Stereo & \checkmark & \checkmark & \textbf{0.091} & \textbf{0.548} & \textbf{3.690} & \textbf{0.181} & \textbf{0.892} & 0.956 & 0.979 \\
\hline
\end{tabular}
}
\smallskip
\caption{Evaluation on KITTI \cite{kitti} odometry sequences from the Eigen split \cite{Eigen} (8691 frames). $^*$ means pre-training on CityScapes \cite{cityscapes}.}
\label{table:eigen}
\end{table*}

\subsection{Ablation studies}

We first run a set of experiments to study the impact of each design choice introduced in our framework. Purposely, we train VOMonodepth in five configurations obtained by: i) directly feeding the depth estimator with VO input (\textit{baseline}) ii) introducing the sparse-autoencoder iii) adding the skip module iv) performing symmetric training on the sparse points v) fine-tuning on $K_o$ and monocular VO.

Table \ref{table:ablation} collects the outcome of this evaluation, carried out on the testing split mentioned above and made of 8691 frames. For comparison, we report the results achieved by Monodepth-ResNet \cite{Godard1}; for all models, we perform the post-processing step introduced in \cite{Godard1}.
We point out that, while the baseline barely outperforms \cite{Godard1} on some metrics, the introduction of the sparse-autoencoder is crucial to boost the effectiveness of our approach. Adding the skip module (+skip) to our architecture enables for a slight improvement on Sq Rel, RMSE and $\delta$ scores.
A major contribution is given by adding the symmetric training (+sym.), optimizing on VO points aligned both on the left and right images.
It is worth noting that the aforementioned three configurations have been trained to leverage VO input from a stereo algorithm, while at deployment such cues comes from a monocular VO algorithm. Although the nature of input VO differs between training and testing, our technique is effective indeed at improving monocular depth estimation.
Finally, by running a fine-tuning (+ft) switching from stereo to monocular VO input allows to a further, major boost on all metrics, leading to the best configuration.

\subsection{Comparison with state-of-the-art}

Table \ref{table:eigen} reports results on the same testing split defined before. We point out once more that, since compliant with the Eigen split \cite{Eigen}, we can compare our proposal to most existing methods. Specifically, we report in the table competitors for whose the source code or trained models are available, self-supervised either using monocular or stereo images. Unfortunately, code and models are not available for \cite{yang2018deep,Pilzer_2019_CVPR}, and hence, we are not able to compare with them.
Methods marked with $^*$ have been pre-trained on CityScapes dataset \cite{cityscapes}, for whose the authors do not provide weights trained on KITTI only.
The table is divided into three portions, from top to bottom: i) monocular supervised, ii) lightweight stereo-supervised and iii) complex stereo-supervised models. Our variants, VOPyD-Net and VOMonodepth, belong respectively to categories ii) and iii).

Starting from compact models, we evaluate PyD-Net \cite{pydnet18} and its VO variant with and without applying the post-processing (PP) step introduced in \cite{Godard1}. However, it is worth observing that since PP requires to forward the input image twice, it adds a non-negligible overhead that is undesirable in case of deployment on embedded systems or when targeting the maximum efficiency \cite{pydnet18}.
VOPyD-Net outperforms PyD-Net by a notable margin on most metrics, except Sq Rel and RMSE. In particular, $\delta<1.25$ receives the highest improvements, \ie 87.5\% pixels are below the threshold versus the 82.5 of PyD-Net. By running the post-processing, VOPyD-Net consistently outperforms it on all metrics by a significant margin.
Moreover, this model also outperforms Monodepth-ResNet on all metrics and 3Net-ResNet on all except $\delta<1.25^2$ and $\delta<1.25^3$, despite the much lower complexity of the network and the lower runtime, as discussed further.

By coupling our strategy with a more complex architecture, as in the case of VOMonodepth-ResNet, we can outperform even MonoResMatch \cite{Tosi_2019_CVPR} which deploys a more accurate architecture and leverages additional supervision from SGM algorithm \cite{hirschmuller2008stereo} at training time. This experiment further proves the effectiveness of our strategy, outperforming state-of-the-art methods for self-supervised monocular depth estimation.

\begin{table}[]
    \centering
    \begin{tabular}{|l|c|c|c|}
    \hline
    Method & PP & TX2 & 2080Ti \\
    & & (Fps) & (Fps) \\
    \hline
    \hline
    PyD-Net \cite{pydnet18} & & 24.22 & 195.34 \\
    VOPyD-Net & & 18.48 & 143.04 \\
    Monodepth - ResNet \cite{Godard1} & & 3.41 & 85.12 \\    
    VOMonodepth - ResNet & & 2.77 & 67.13 \\
    \hline
    VOPyD-Net & \checkmark & 8.35 & 100.11 \\
    Monodepth - ResNet \cite{Godard1} & \checkmark & 2.24  & 
62.21 \\
    3Net-ResNet\cite{3net18} & \checkmark & 2.10 & 49.24 \\
    VOMonodepth - ResNet & \checkmark & 1.70 & 41.84 \\
    MonoResMatch \cite{Tosi_2019_CVPR} & \checkmark & 1.23 & 29.79 \\
    \hline
    \end{tabular}
    \caption{Runtime comparison (averaged over 200 frames) between previous models and VO
variants on Jetson TX2 and 2080Ti.}
    \label{tab:fps}
\end{table}

\begin{figure*}
    \centering
    \renewcommand{\tabcolsep}{0.5pt}
    \begin{tabular}{cccc}

        \scriptsize
        \rotatebox[origin=l]{90}{Reference} &
        \includegraphics[width=0.28\textwidth]{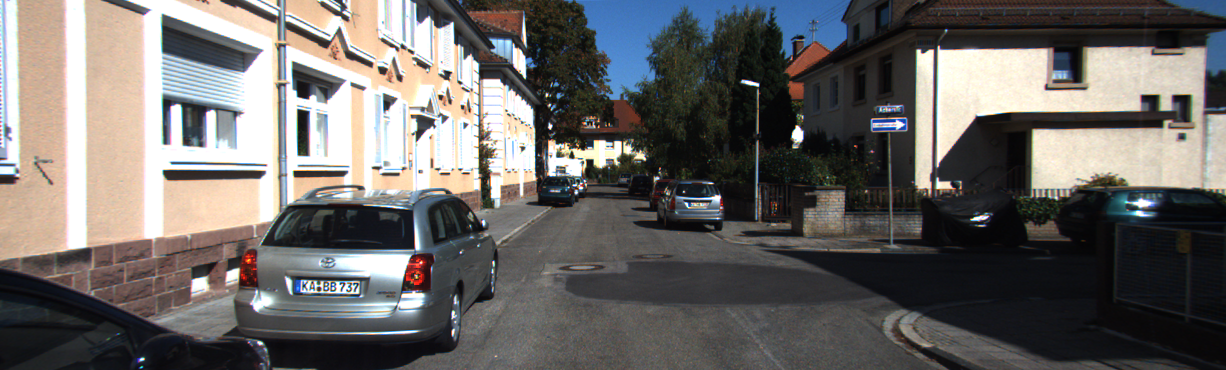} &
        \includegraphics[width=0.28\textwidth]{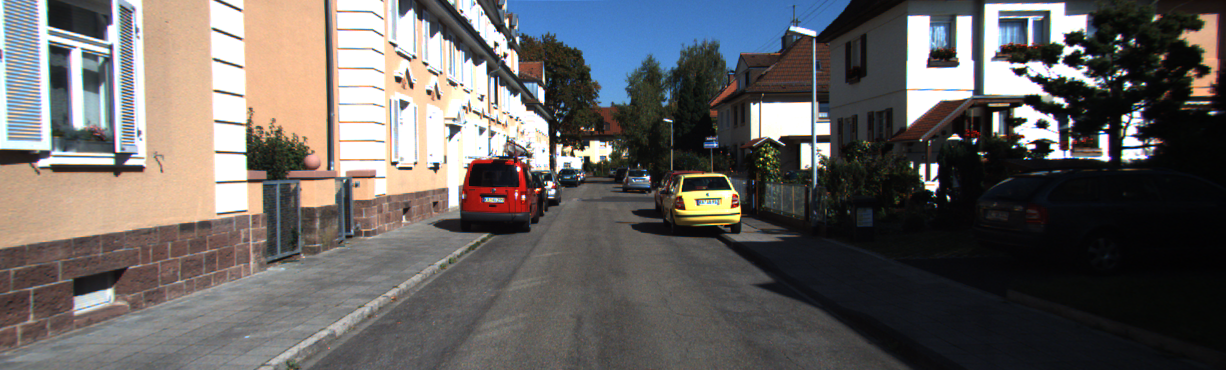} &
        \includegraphics[width=0.28\textwidth]{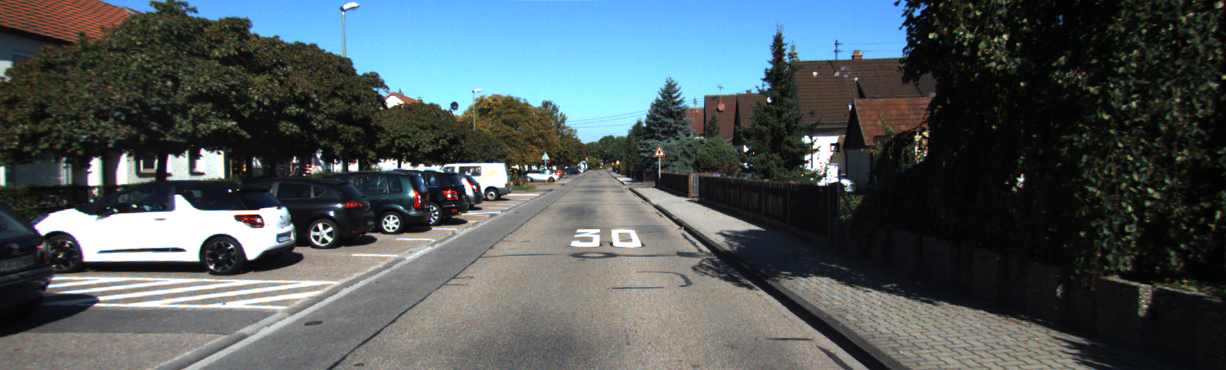}
        \\
        \scriptsize
        \rotatebox[origin=l]{90}{SD} &
        \includegraphics[width=0.28\textwidth]{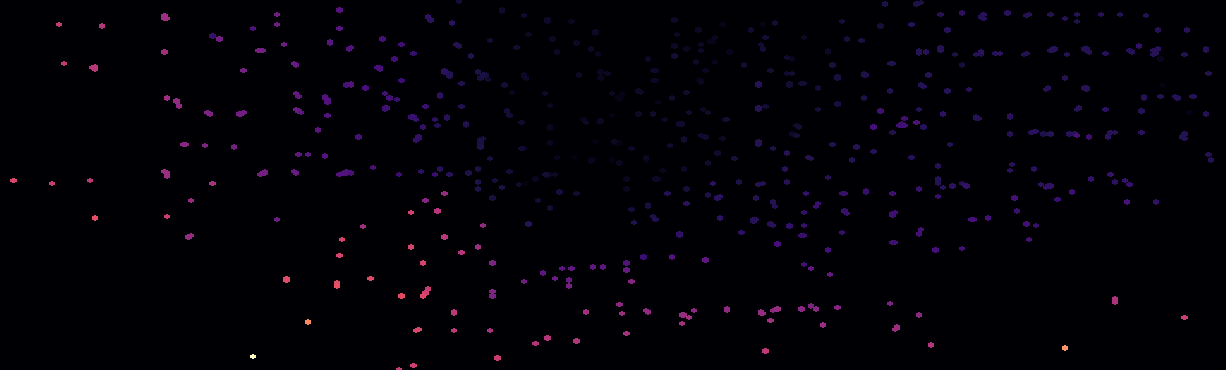} &
        \includegraphics[width=0.28\textwidth]{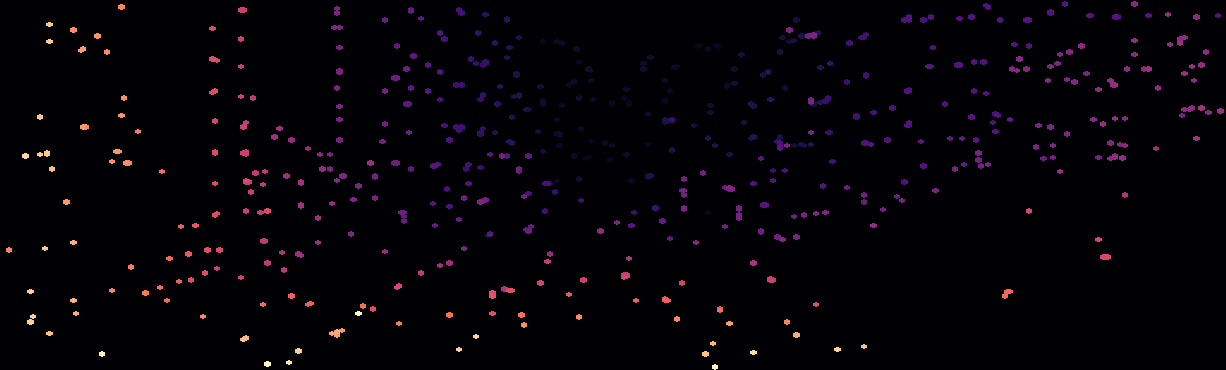} &
        \includegraphics[width=0.28\textwidth]{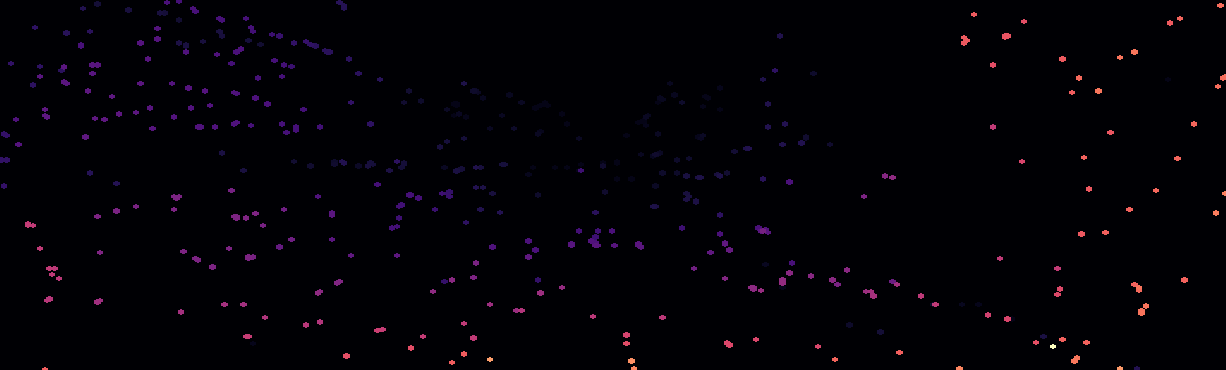}
        \\
        \scriptsize
        \rotatebox[origin=l]{90}{PyD-Net\cite{pydnet18}} &
        \includegraphics[width=0.28\textwidth]{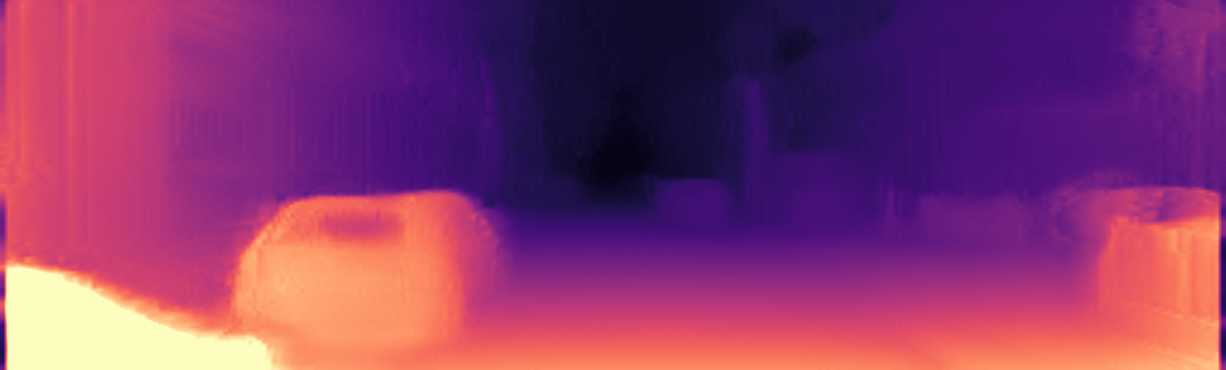} &
        \includegraphics[width=0.28\textwidth]{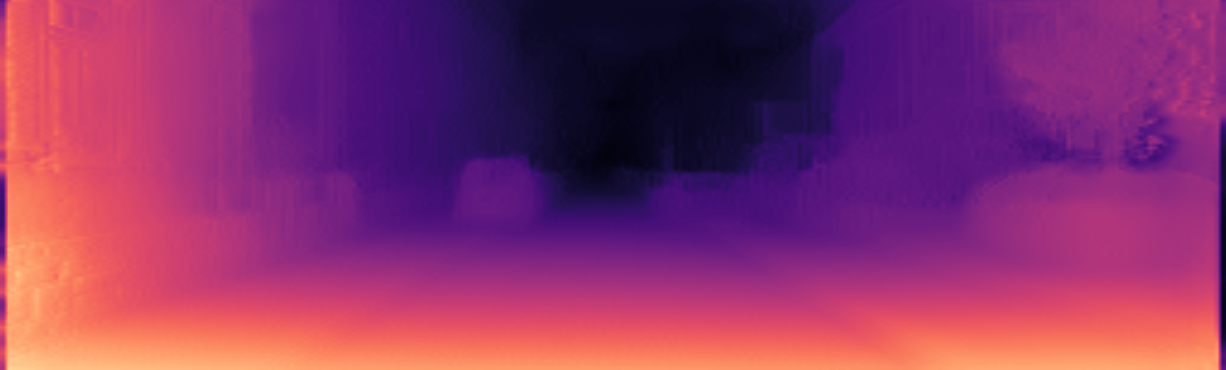} &
        \includegraphics[width=0.28\textwidth]{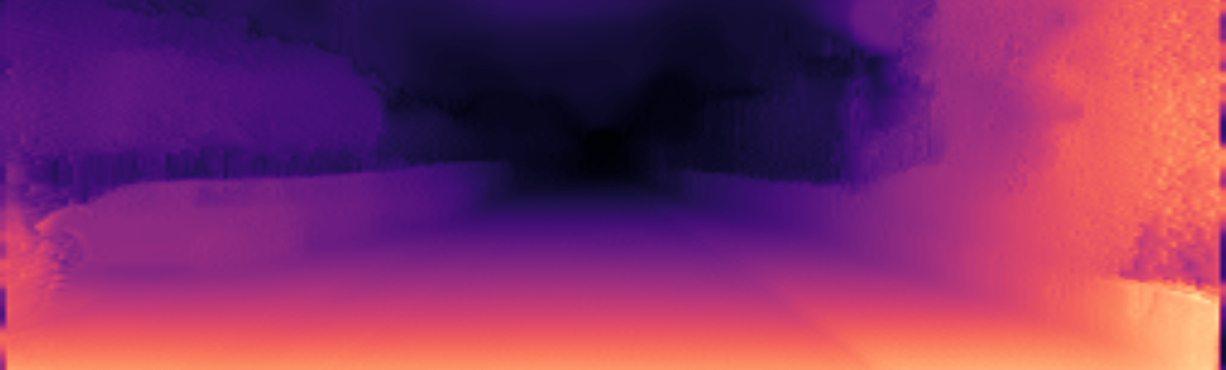}
        \\
        \scriptsize
        \rotatebox[origin=l]{90}{VOPyD-Net} &
        \includegraphics[width=0.28\textwidth]{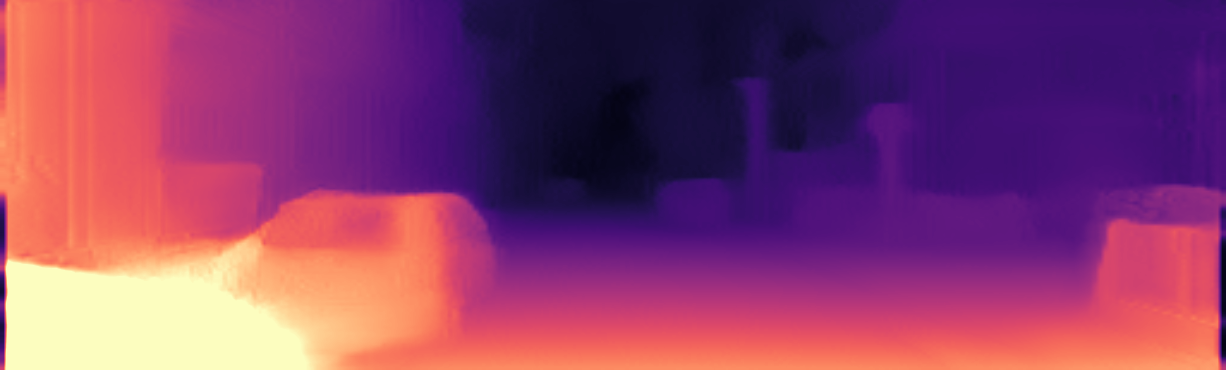} &
        \includegraphics[width=0.28\textwidth]{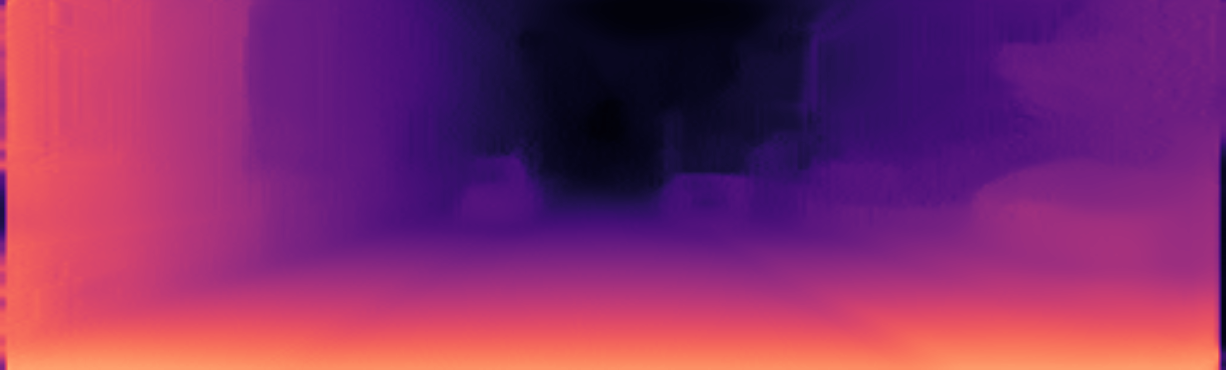} &
        \includegraphics[width=0.28\textwidth]{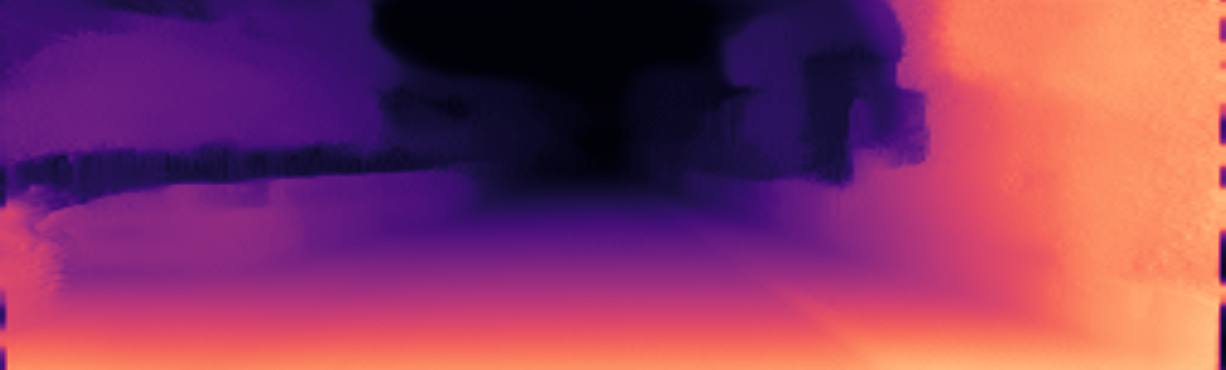}
        \\
        \scriptsize
        \rotatebox[origin=l]{90}{Monodepth\cite{Godard1}} &
        \includegraphics[width=0.28\textwidth]{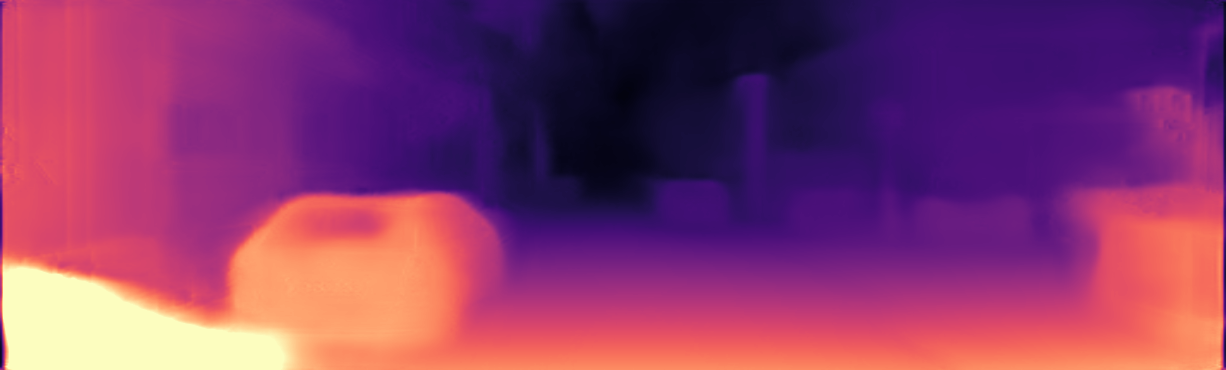} &
        \includegraphics[width=0.28\textwidth]{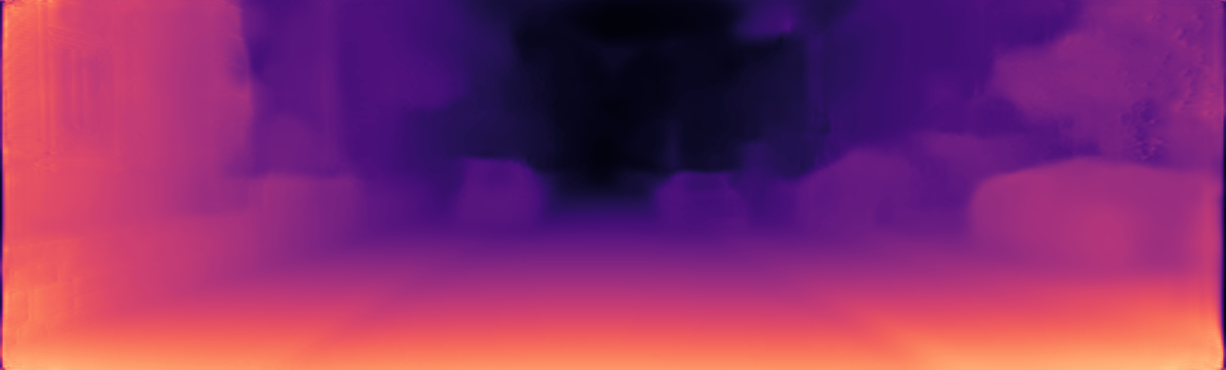} &
        \includegraphics[width=0.28\textwidth]{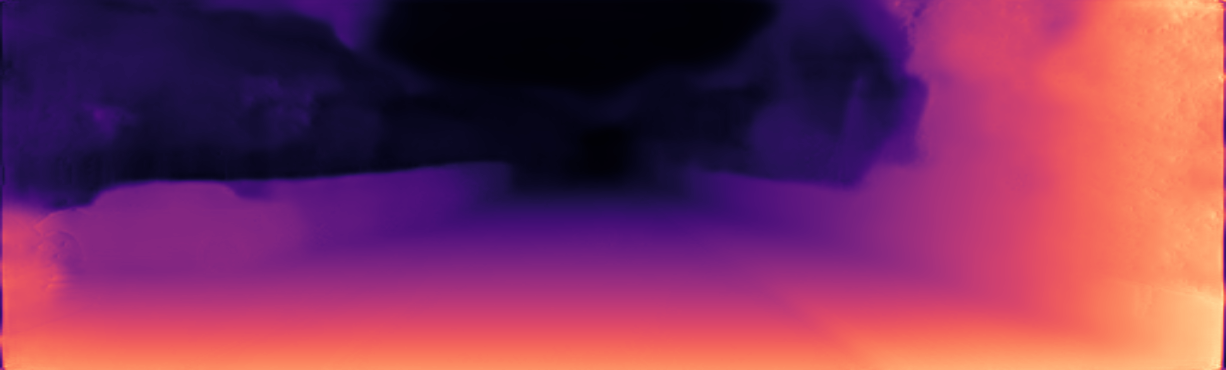}
        \\ 
        \scriptsize
        \rotatebox[origin=l]{90}{VOMonodepth} &
        \includegraphics[width=0.28\textwidth]{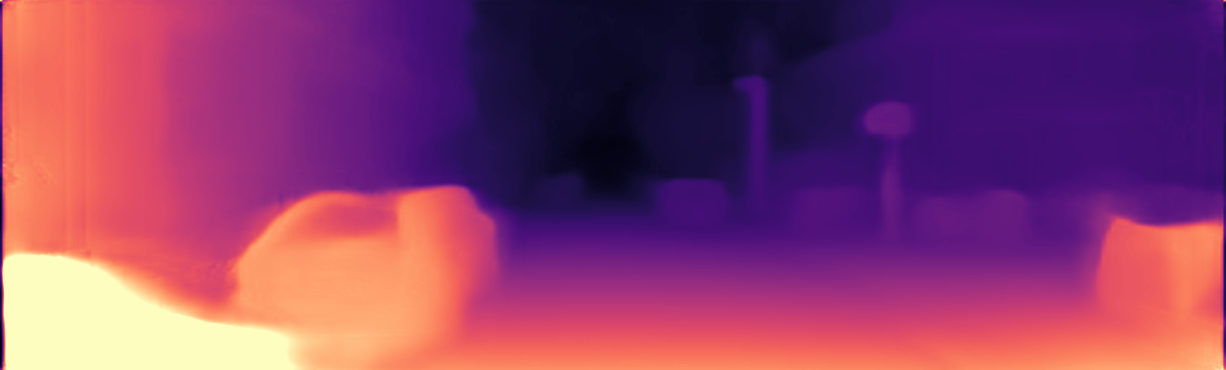} &
        \includegraphics[width=0.28\textwidth]{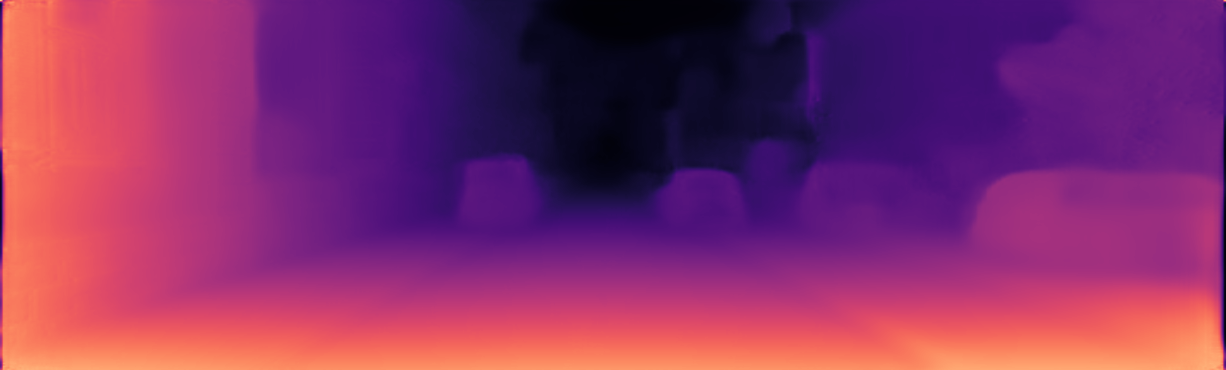} &
        \includegraphics[width=0.28\textwidth]{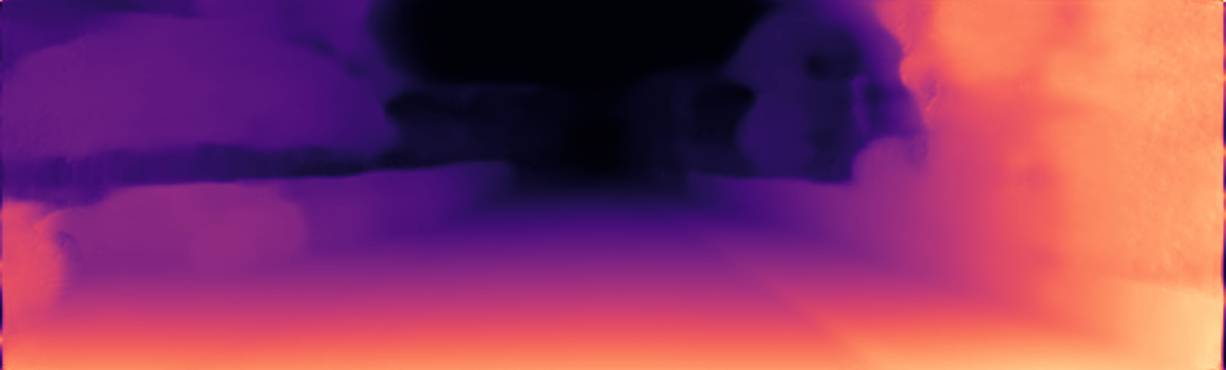}
        \\
        \scriptsize
        \rotatebox[origin=l]{90}{LiDAR} &
        \includegraphics[width=0.28\textwidth]{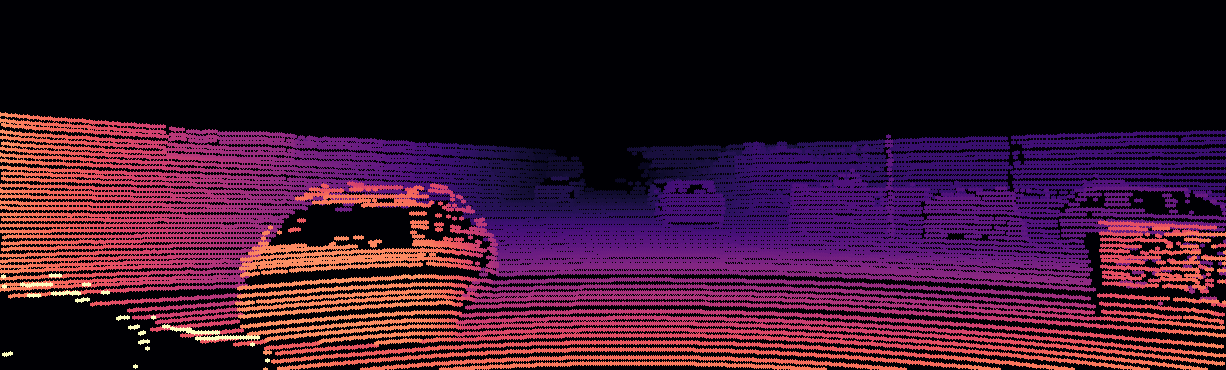} &
        \includegraphics[width=0.28\textwidth]{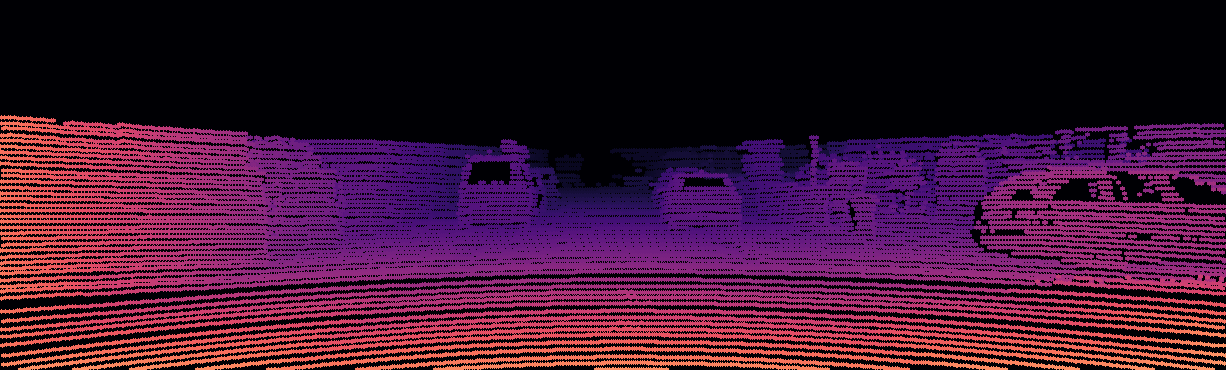} &
        \includegraphics[width=0.28\textwidth]{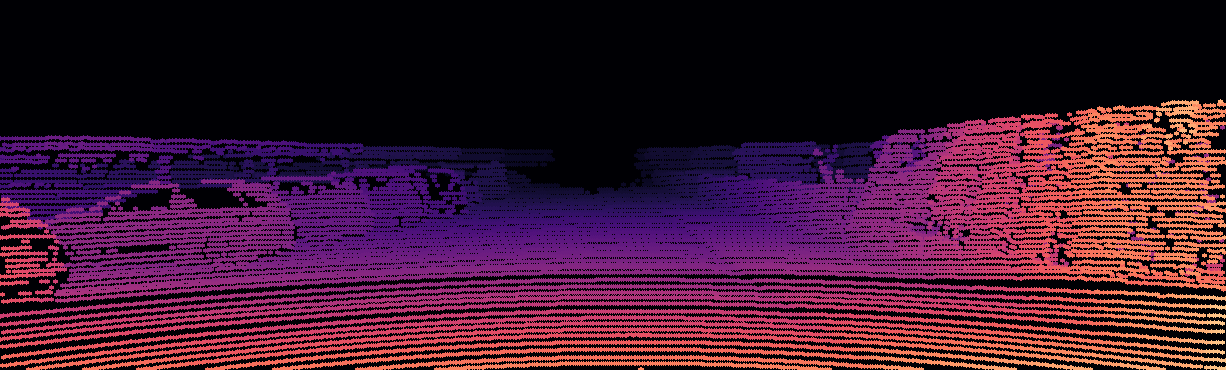}
        \\        
    \end{tabular}
    \normalsize
    \caption{Qualitative results on KITTI dataset. On each column, from top to bottom: reference image, sparse VO points, depth map outputs from PyD-Net \cite{pydnet18}, VOPyD-Net, Monodepth \cite{Godard1}, VOMonodepth and LiDAR points used for evaluation. \label{fig:qualitatives}}
\end{figure*}

\begin{figure}
    \centering
    \renewcommand{\tabcolsep}{0.5pt}
    \begin{tabular}{cc}
        \begin{overpic}[width=0.21\textwidth, height=0.08\textwidth]{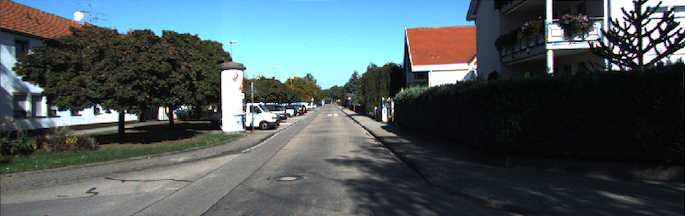}
        \put (2,24) {$\displaystyle\textcolor{white}{\textbf{(a)}}$}
        \end{overpic} 
        &
        \begin{overpic}[width=0.21\textwidth, height=0.08\textwidth]{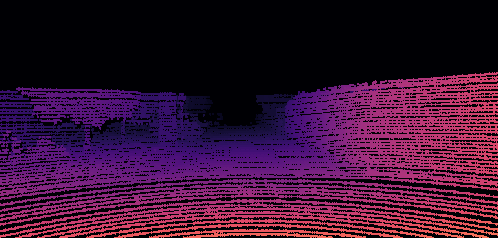}
        \put (2,24) {$\displaystyle\textcolor{white}{\textbf{(b)}}$}
        \end{overpic} 
        \\
        \begin{overpic}[width=0.21\textwidth, height=0.08\textwidth]{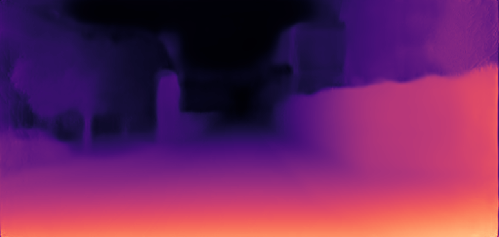}
        \put (2,24) {$\displaystyle\textcolor{white}{\textbf{(c)}}$}
        \end{overpic} 
        &
        \begin{overpic}[width=0.21\textwidth, height=0.08\textwidth]{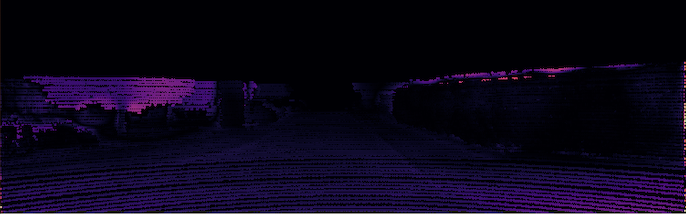}
        \put (2,24) {$\displaystyle\textcolor{white}{\textbf{(d)}}$}
        \end{overpic} 
        \\
        \begin{overpic}[width=0.21\textwidth, height=0.08\textwidth]{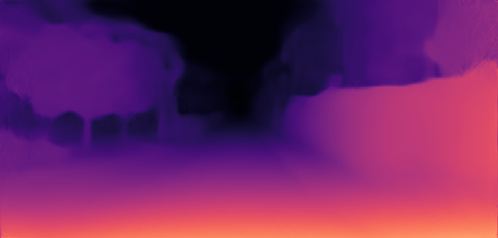}
        \put (2,24) {$\displaystyle\textcolor{white}{\textbf{(e)}}$}
        \end{overpic} 
        &
        \begin{overpic}[width=0.21\textwidth, height=0.08\textwidth]{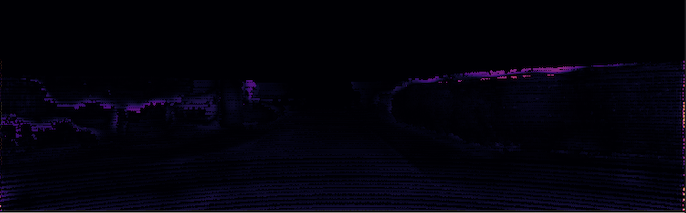}
        \put (2,24) {$\displaystyle\textcolor{white}{\textbf{(f)}}$}
        \end{overpic} 
        \\        
    \end{tabular}
    \caption{Qualitative comparison on KITTI dataset. (a) input image, (b) LiDAR points, depth and error maps respectively by Monodepth (c,d) and VOMonodepth (e,f).  
    \label{fig:qualitatives2}}
\end{figure}

\subsection{Runtime analysis}

We benchmark the performance of our VO variants and traditional models on two very different hardware platforms: an NVIDIA 2080 Ti GPU, having 250W of power consumption, and a Jetson TX2 module with a maximum average consumption of about 15W. The latter device represents an appealing platform for a wide range of applications. In all experiments, the TX2 board was configured for maximum performance.
Table \ref{tab:fps} collects the results of this analysis, comparing Monodepth and PyD-Net with their VO counterparts. We report Fps both enabling and disabling post-processing.

Focusing on the TX2 platform, we can notice how the difference between PyD-Net and VOPyD-Net is about 30\%, running respectively at 24 and 18 Fps, if post-processing is disabled, about $7\times$ and $5\times$ faster than \cite{Godard1} (3.41). A similar overhead, about 20\%, is introduced comparing VOMonodepth-ResNet to Monodepth-ResNet.
Enabling PP, VOPyD-Net still runs at more than 8 Fps, but it produces better results on each metrics (see Table \ref{table:eigen}) compared to Monodepth-ResNet, despite running more than $3\times$ faster. It also outperforms on most metrics 3Net-ResNet, running almost $4\times$ faster. VOMonodepth achieves much higher accuracy at the cost of a further drop in speed (below 2 Fps), but still $40\%$ faster than MonoResMatch. 

Switching to NVIDIA 2080Ti GPU, a similar overhead between each model and its VO variant can be noticed. Even enabling post-processing, VOPyD-Net still runs at about 100 Fps versus the 62 and 49  by Monodepth-ResNet and 3Net-ResNet, while VOMonodepth-ResNet reaches about 40 Fps versus the 29 of MonoResMatch, making the former the best choice when high-end GPUs are available for deployment thanks to its superior accuracy.

Finally, Figure \ref{fig:qualitatives} shows some qualitative examples of depth map inferred by both PyD-Net and Monodepth as well as their VO counterparts. We can notice how sparse inputs improve, for instance, estimates on thin structures (left column). Figure \ref{fig:qualitatives2} shows a comparison between Monodepth \cite{Godard1} and VOMonodepth, reporting error maps (d), (f) to highlight how the former fails at estimating the depth for trees on the left, whereas our method is successful.
\section{Conclusion}
\label{conclusions}

In this paper, we have proposed a novel framework that takes into account prior knowledge to improve monocular depth estimation. We have introduced a geometrical prior obtained by estimating the movement of the camera, as it commonly happens in an autonomous navigation scenario. Our network is able to leverage on the sparse 3D measurements of a VO algorithm to improve depth accuracy.
Extensive experimental results on the KITTI dataset prove that our framework: i) outperforms existing models for self-supervised depth estimation and ii) it is practical and couples with complex and compact models, allowing for accurate, real-time monocular depth estimation on high-end GPUs as well as on embedded systems.

\textbf{Acknowledgements.} We thank Zenuity for providing their VO pipeline for our experiments. We gratefully acknowledge the support of NVIDIA Corporation with the donation of the Titan Xp GPU used for this research. 

{\small
\bibliographystyle{ieee}
\bibliography{bibl}
}

\end{document}